\documentclass[twoside]{article}

\usepackage{PRIMEarxiv}

\usepackage[utf8]{inputenc} 
\usepackage[T1]{fontenc}    
\usepackage{hyperref}       
\usepackage{url}            
\usepackage{booktabs}       
\usepackage{amsfonts}       
\usepackage{nicefrac}       
\usepackage{microtype}      
\usepackage{lipsum}
\usepackage{fancyhdr}       
\usepackage{graphicx}       
\graphicspath{{media/}}     
\usepackage{caption}
\usepackage[export]{adjustbox}
\usepackage[table,xcdraw,dvipsnames]{xcolor}
\usepackage{float}
\usepackage{setspace} 
\usepackage{lineno}
\usepackage{rotating}
\usepackage{graphicx}
\usepackage{subfigure}
\usepackage{multirow}
\usepackage{amsmath}
\usepackage{makecell}
\usepackage[flushleft]{threeparttable}

\usepackage{soul}
\usepackage[toc,page]{appendix}

\usepackage{lmodern,textcomp}

\title{Harmful algal bloom forecasting. A comparison between stream and batch learning
\thanks{\textit{\underline{Citation}}: 
\textbf{}}
}
\author{
  Andres Molares-Ulloa, Daniel Rivero, Enrique Fernandez-Blanco \\
  Universidade da Coruña, Department of Computer Science and Information Technology, \\
  Faculty of Computer Science, 15071, A Coruña, Spain \\
  Centro de investigación CITIC, Department of Computer Science and Information Technology, \\ 
  University of A Coruña, 15071, A Coruña, Spain \\
  \texttt{andres.molares@udc.es} \\
   \And
  Xosé A. Padin \\
Department of Oceanography, Instituto de Investigaciones Mariñas (IIM-CSIC), Vigo, Spain \\
   \And
  Elisabet Rocruz, Rita Nolasco, Jesús Dubert \\
Department of Oceanography, Instituto de Investigaciones Mariñas (IIM-CSIC), Vigo, Spain \\
Department of Physics, Centro de Estudos do Ambiente e do Mar, University of Aveiro, Aveiro, Portugal\\
}

\begin{document}
\maketitle

\begin{abstract}
Diarrhetic Shellfish Poisoning (DSP) is a global health threat arising from shellfish contaminated with toxins produced by dinoflagellates. The condition, with its widespread incidence, high morbidity rate, and persistent shellfish toxicity, poses risks to public health and the shellfish industry. High biomass of toxin-producing algae such as DSP are known as Harmful Algal Blooms (HABs). Monitoring and forecasting systems are crucial for mitigating HABs impact. The study of phytoplankton blooms and their causes is challenging, prompting the exploration of advanced machine learning techniques. Predicting harmful algal blooms involves a time-series-based problem with a strong historical seasonal component, however, recent anomalies due to changes in meteorological and oceanographic events have been observed. Stream Learning stands out as one of the most promising approaches for addressing time-series-based problems with concept drifts. However, its efficacy in predicting HABs remains unproven and needs to be tested in comparison with Batch Learning. Historical data availability is a critical point in developing predictive systems. In oceanography, the available data collection can have some constrains and limitations, which has led to exploring new tools to obtain more exhaustive time series. Ocean hydrodynamic models, such as CROCO, has proven its ability to reproduce the ocean conditions with a high temporal and spatial frequency. Model explainability is increasingly crucial for decision acceptance, particularly in the context of alarm systems like HAB forecasting. In this study, a machine learning workflow for predicting the number of cells of a toxic dinoflagellate, \textit{Dinophysis acuminata}, was developed with several key advancements. Seven machine learning algorithms were compared within two learning paradigms. Notably, the output data from CROCO, the ocean hydrodynamic model, was employed as the primary dataset, palliating the limitation of time-continuous historical data. This study highlights the value of models interpretability, fair models comparison methodology, and the incorporation of Stream Learning models. The model DoME, with an average $R^2$ of 0.77 in the 3-day-ahead prediction, emerged as the most effective and interpretable predictor, outperforming the other algorithms. 
\end{abstract}


\section{Introduction}
\label{S:1}

Diarrhetic shellfish poisoning (DSP) is a gastrointestinal disease resulting from the consumption of shellfish contaminated with toxins produced by dinoflagellates. Outbreaks of DSP have been reported worldwide, threatening human health, aquaculture production and originates high economic losses to coastal related activities, as tourism \cite{GRATTAN20162}.
High biomass of algal producing toxin such as DSP are known as Harmful Algal Blooms (HABs). Seasonal outbreaks of DSP generated on the Galician coast (northwestern Spain) have been associated with recurrent blooms of \textit{Dinophysis acuminata} \cite{diaz2013climate, Velo-Suarez2014, Ruiz-Villarreal2016, Blanco2019, Diaz2019b, Brown2020, Velasco-Senovilla2023}. HABs constitute a natural phenomenon that cannot be avoided
and the study of the ecological and oceanographic mechanisms underlying its occurrence is a challenging field \cite{LEE2003179}. Different approaches have been used, from the study of their life cycles \cite{Reguera2024}, their advection through individual-based models \cite{Li2014, Moita2016} to their prediction through the used of machine learning techniques \cite{VeloSurez2007ArtificialNN,Silva2023}. Associated to the complex relations between many biological variables involved in HABs forecasting and the environmental (oceanographic and meteorological) parameters, various machine learning techniques have been studied in recent years, achieving better results than more classical predictive techniques \cite{cruz2021review}. Methods such as Multi-layer Perceptron (MLP) \cite{1553742, GUO2020111731}, support vector machines (SVM) \cite{RIBEIRO200886, 7043865} or tree-based techniques such as Random Forest (RF) \cite{DEROT2020101906, HARLEY2020101918} or XGBoost \cite{rs13193863, MOLARESULLOA2023107988} are the most popular.

Another factor that is becoming increasingly important in the development of forecasting models is their explainability \cite{xai}. Tree-based models tend to be the most widely used to fulfil this requirement \cite{YAN2024169253}. This should be a key factor in future work, as the creation of an alarm system, such as the HAB forecast, should allow the identification of the main forcings of algal blooms to improve the understanding and reliability of these predictive tools. As a consequence, this would also lead to a better explanation and acceptance of the decision taken by the responsible managers of the aquaculture industry.

Forecasting harmful algal blooms is a prediction problem based on time series analysis. Although these time series historically have a strong seasonal component, there are environmental evidences that show significant anomalies in the occurrence of these episodes associated to changes in meteorological and oceanographic conditions \cite{diaz2013climate,BOIVINRIOUX2022102183}. One of the paradigms that have shown the best results when dealing with this type of problem is Stream Learning. This is a machine learning paradigm in which the models learn incrementally from a stream of data points in real time \cite{HOI2021249}. This paradigm has been applied in successfully solving numerous time series regression problems \cite{8616838, benjamin2015real, wang2016dynamically}.

The set of historical data available is one of the most critical points when developing this type of predictive systems. In oceanography, sample collection may have some limitations to obtain a continuous time series, such as adverse weather conditions, equipment malfunctions, availability of research vessels and sufficient funding from the responsible institutions to meet these structural costs. In this sense, the use of hydrodynamic models have been widely incorporated to better understand HABs dynamic \cite{GEOHAB2011, Ruiz-Villarreal2016}. To this end, ocean hydrodynamic models, such as the Coastal and Regional Ocean COmmunity model (CROCO; \cite{francis_auclair_2022_7415343}), capable of provide quality trustable environmental output data with saving the output data with a high temporal frequency, may improve HAB prediction through machine learning techniques, which is yet to prove.


Therefore, in this study, we have developed a machine learning model capable of predicting the number of cells of the toxic dinoflagellate \textit{Dinophysis acuminata}, the main responsible for DSP closures in aquaculture production in Galicia. This has been achieved through three main advances: (1) The use of 7 machine learning algorithms enclosed in 2 learning paradigms: Stream Learning (k-Nearest Neighbor (kNN), Hoeffding Tree Regressor (HTR) and Hoeffding Adaptive Tree Regressor (HATR)) and Batch Learning (kNN, Support Vector Regressor (SVR), Multi-layer Perceptron (MLP), Random Forest (RF) and DoME). While Batch Learning has been used in HABs forecasting, Stream Learning has not yet been tested in this field. kNN, SVR, and MLP were chosen as benchmarks because they are commonly used in these studies \cite{MOLARESULLOA2022106956}. DoME \cite{RIVERO2022116712}, RF \cite{YAN2024169253}, HTR, HATR \cite{IKONOMOVSKA2015458} and kNN were chosen for their high explainability. (2) The development of a methodology that allows a fair comparison between the models previously used in the literature and the models proposed in this work. (3) The use of the output data from CROCO model as the main part of the dataset and their combination with machine learning techniques. This combination has not yet been tested to the authors knowledge.

\section{Materials and Methods} 
\label{sec:mm}
\subsection{Dataset and its construction}

\begin{figure}[ht]
  \centering
  \includegraphics[trim = 1000mm 0mm 1000mm 0mm, clip,scale=0.27]{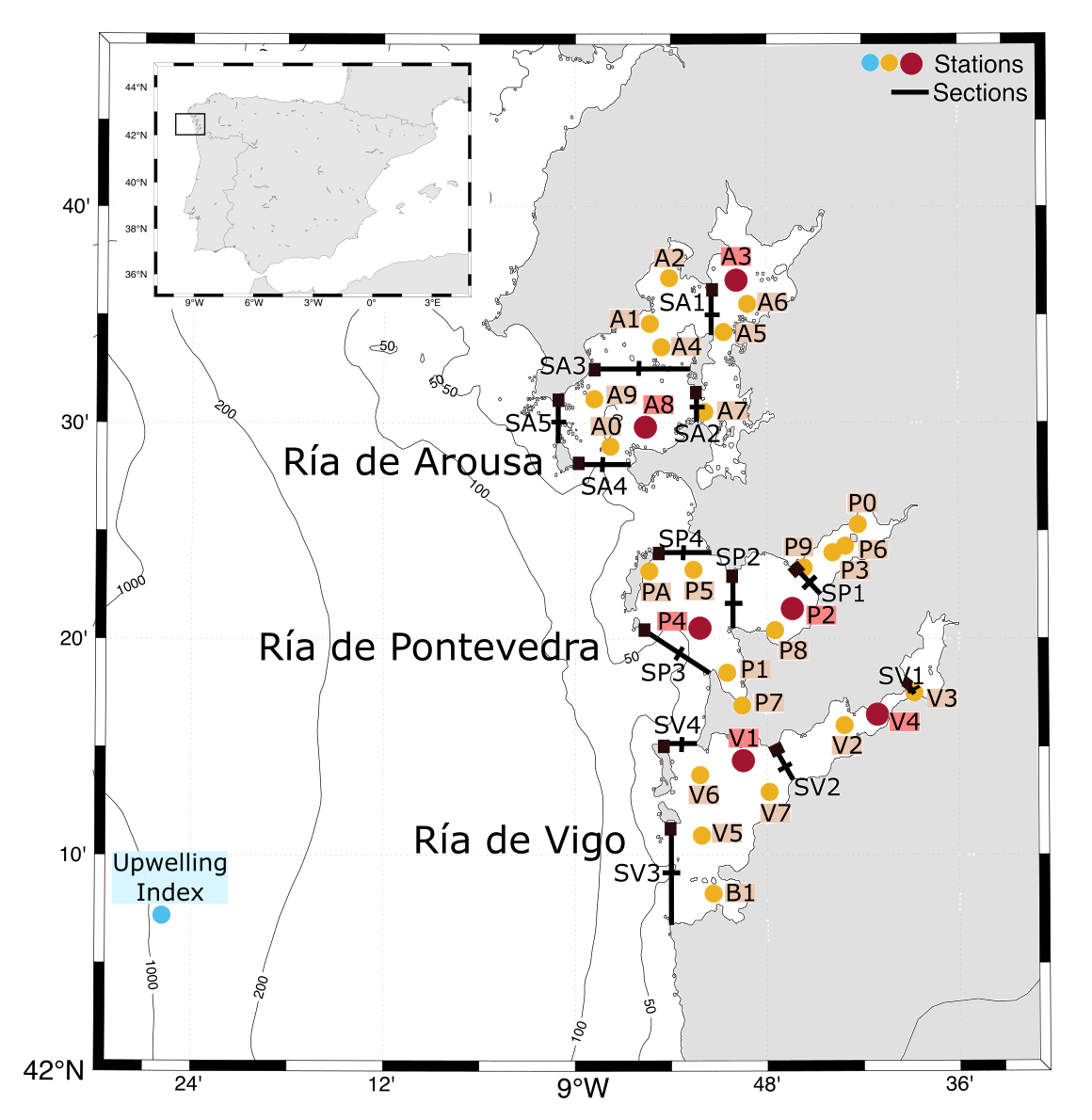}
  \caption{Map of the Galician Rias Baixas, Arousa, Pontevedra and Vigo with the INTECMAR stations (coloured dots within the Rias), Upwelling Index station (blue dot at the shelf) and sections (black lines). Red dots represent the stations where the prediction was done, orange dots the stations which input data was used for the prediction. In the section, the first half goes from the black square to the central cross-line and the second half from here to the end.}
  \label{fig:mapa}
\end{figure}

The input data in this study consisted on the spatiotemporal distributions of biological and physical variables for a period of seven years (2013-2019). The biological data included Chlorophyll-a and \textit{D. acuminata} counting cells. Daily surface Chlorophyll-a (mg/m$^{3}$) obtained from SeaWIFS, MERIS, MODIS-A,
MODIS-T, VIIRS-SNPP \& JPPS1, OLCI-S3A \& S3B satellites using the multi sensor MY, was downloaded from Copernicus\footnote{\url{https://doi.org/10.48670/moi-00289}}. Abundance of \textit{D. acuminata} (cells/L) cells was provided by ``Instituto Tecnolóxico para o Control do Medio Mariño de Galicia'' (INTECMAR). Since the raw data are sampled weekly, not always the same day of the week, and some values are missing, we applied linear interpolation to obtain daily values. These biological variables were obtained at the mooring stations (coloured dots within the Rias in Figure \ref{fig:mapa}), which are defined by INTECMAR in its sampling programme.

The physical variables, such as daily temperature, salinity and 3D velocity components, were generated from an 3D ocean hydrodynamic model, CROCO \cite{francis_auclair_2022_7415343}. It is built upon the ROMS-AGRIF \cite{Shchepetkin2005}. CROCO is a three-dimensional, split-explicit, free-surface and a terrain-following s-coordinate numerical model and has the capability of resolving very fine spatial scales, through two-way nesting domains. The model configuration used for this study had three two-way nested domains, with the smallest size grid covering the Galician Rias Baixas, and with a resolution of ca. 180 m. CROCO model configuration implemented as input data: meteorological and ocean forcings, freshwater riverine flows and tides. Further details of this configuration are described in \cite{ElisabetCruz2021}. 

The previously described physical and biological variables were distributed at the three southernmost Rias Baixas (Figure \ref{fig:mapa}), Ría de Arousa, Pontevedra and Vigo, along the mooring stations (coloured dots within the Rias in Figure \ref{fig:mapa}) and sections (black lines in Figure \ref{fig:mapa}). At the stations (coloured dots in Figure \ref{fig:mapa}), the mean and standard deviation of temperature, salinity, and zonal and meridional velocity components computed by CROCO model were recorded at surface (0-15 m depth) and at bottom, when station depth was higher than 15 m. Another physical variable recorded at the stations was the maximum value of the Brunt-Väisälä (BV) frequency (s$^{-2}$) and the depth of the water column at which this maximum was located. It was calculated from the density profiles using the model output variables. The BV frequency measures the stability of the water column and the absolute maximum value represents a change in the stratification of the water column. As an approach of the ocean-Rias and inner-outer Rias exchange, physical variables were also recorded along sections (black lines in Figure \ref{fig:mapa}). These variables were the mean and standard deviation of the temperature, salinity, and the 3D velocity components.  In the sections, the mean and standard deviation were calculated at the surface (0-15 m depth) and at the bottom (>15 m depth), and were also calculated in each half of the sections. First half of the sections is indicated between the black square and the cross-line and the second from here to the end of the section (see Figure \ref{fig:mapa}).

An additional physical variable incorporated on the machine learning models was the Upwelling Index (UI; m$^{3}$ s$^{-1}$ km$^{-1}$). Positive values of UI \cite{Bakun1973} indicate a predominance of northerly wind components, which induces the rise of cooler, nutrient-rich subsurface waters near the surface, while surface water moves offshore (upwelling conditions). Southerly winds, with negative values of UI, have the opposite effect, that is the transport of surface waters towards the coast (downwelling conditions). UI values were computed from the National Centers for Environmental Prediction (NCEP) wind reanalysis data at 42.12$^{\circ}$N and 9.43$^{\circ}$W (blue dot in Figure \ref{fig:mapa}). The number of the month by applying a sine function was also included in the machine learning models.

The forecasting of the concentration of \textit{D. acuminata} cells was performed at two stations representative of the outer and inner part (red dots in Figure \ref{fig:mapa}) at each Ría. In the Ría de Arousa, stations A8 and A3 were selected, in the Ría de Pontevedra, P4 and P2, and in the Ría de Vigo, V1 and V4. To predict \textit{D. acuminata} cells concentration at these stations, in addition to including the month, UI at the shelf and the physical and biological data at the station to be predicted, data from the surrounding stations (orange dots in Figure \ref{fig:mapa}) and sections were also incorporated. The selected stations and sections are detailed in Table \ref{tab:StaSec}.

\begin{table}[ht]
\centering
\caption{Stations and sections used to predict \textit{D. acuminata} cells at an outer and inner station at Ría de Arousa, Pontevedra and Vigo. Outer stations and sections are indicated with an asterisk.}
\label{tab:StaSec}
\begin{tabular}{ccc}
\hline
 & \multicolumn{2}{c}{\textbf{Input data}} \\ \cline{2-3} 
\multirow{-2}{*}{\textbf{Predict}} & \textbf{Stations} & \textbf{Sections} \\ \hline
{\textbf{A8$^{*}$}} & A8$^{*}$, A9$^{*}$, A0$^{*}$, A7, A4 & SA5$^{*}$, SA4$^{*}$, SA3, SA2 \\ \hline
{\textbf{A3}} & A3, A6, A5, A2, A1 & SA2, SA1 \\ \hline
{\textbf{P4$^{*}$}} & P4$^{*}$, P5$^{*}$, PA$^{*}$, P1$^{*}$, P2 & SP4$^{*}$, SP3$^{*}$, SP2 \\ \hline
{\textbf{P2}} & P2$^{*}$, P9, P8 & SP2, SP1 \\ \hline
{\textbf{V1$^{*}$}} & V1$^{*}$, V7$^{*}$, V6$^{*}$, V5$^{*}$, V2 & SV4$^{*}$, SV3$^{*}$, SV2 \\ \hline
{\textbf{V4}} & V4, V3, V2 & SV2, SV1 \\ \hline
\end{tabular}
\end{table}

\subsection{Learning paradigms}

In this study, two machine learning paradigms for prediction of \textit{D. acuminata} cells have been compared and the main differences between the two approaches are presented and discussed. 

\subsubsection{Batch Learning}

The conventional machine learning paradigm usually operates in Batch Learning (BL) mode \cite{russell2010artificial}. In this approach, a predefined set of training data is provided in advance to train a model using a specific learning algorithm. This paradigm requires the availability of the entire training data set prior to the learning process, and training usually occurs in an offline environment due to the high computational cost associated with training. These types of methods offer the best performance for problems with high data redundancy. However, traditional Batch Learning methods have significant drawbacks, such as their low efficiency in terms of time and space costs and their poor scalability for large-scale applications. This is mainly because these methods often require the model to be retrained from scratch when new training data is introduced.

\subsubsection{Stream Learning}

Stream Learning (SL), also called Online learning, represents a machine learning paradigm adapted to the use of time series \cite{HOI2021249}. In this paradigm, the model is trained with individual samples. This contrasts radically with the conventional approach of training a model with a whole batch of data at a time. Concept drift is a phenomenon in which the underlying patterns in the data change over time. While this poses a challenge for both offline and online models, the strength of online models lies in their ability to adapt to conceptual drift. Their continuous learning capability allows them to adjust smoothly to changing patterns, unlike batch models that require complete retraining. Another major difference lies in model evaluation. Instead of the typical cross-validation used in batch models, where the training data set is divided into training and testing sets, online models evaluation mirrors the real-world data generation process. By processing data in the same order as it arrives, mimicking a production scenario, online models evaluation provides a more actual representation of model performance compared to cross-validation. Consequently, online learning algorithms demonstrate increased efficiency and scalability, especially in real-world data analytics applications characterised by substantial data volume and high data arrival velocity. 

\subsection{Base Models}

When comparing these two learning paradigms, a total of 7 machine learning models have been used. Some of these models have been designed for the Stream Learning paradigm such as Hoeffding Trees while others can be implemented using both paradigms such as k-Nearest Neighbor. We have briefly explained these base models below.

\subsubsection{k-Nearest Neighbors}

The k-Nearest Neighbor (kNN) classifier represents an unsupervised machine learning method utilized to classify unlabeled observations by associating them with the class of the most similar labeled examples \cite{lantz2013machine}. Features from both the training and test datasets are gathered for these observations. The prevalent metric employed in these calculations is the Euclidean distance. Another critical aspect is the parameter $k$, which determines the number of neighbors selected for the kNN algorithm. The judicious selection of $k$ holds substantial influence over the diagnostic performance of the kNN algorithm, underscoring the importance of choosing an appropriate value to optimize classification accuracy and effectiveness. kNN is recognized as a robust learner, meaning that with a minor variation in the training sample for two KNN models, their predictions will exhibit a considerable degree of similarity.

\subsubsection{Hoeffding Tree Regressor}

Hoeffding Tree Regressor (HTR) is a supervised machine learning model that relies on Decision Trees (DT). DT resembles an inverted tree, where each internal node represents a decision based on a specific feature, each branch corresponds to the outcome of that decision, and each leaf node represents the final decision or the assigned class label. HTR initiate their construction with only one node (leaf). The tree gradually expands as the data undergoes processing \cite{domingos2000mining}. Like FIMT-DD \cite{ikonomovska2011learning}, HTR makes split decisions based on variance information, and the aggregation at the leaves can be achieved either through a linear model, such as a perceptron, or by considering the mean target values of examples reaching the leaf. However, it's important to note that unlike FIMT-DD, HTR lacks mechanisms for adapting to concept drift. This absence means that HTR may struggle with non-stationary distributions when used independently. Nevertheless, this limitation is offset by an improvement in computational resource efficiency compared to FIMT-DD. This aspect makes HTR a favorable choice in scenarios where computational resources are a crucial consideration, even though it may not handle concept drift as effectively as its counterpart.

\subsubsection{Hoeffding Adaptive Tree Regressor}

The Hoeffding Adaptive Tree Regressor (HATR) algorithm builds decision trees using the data stream and adjusts the tree following the examination of each example \cite{10.1007/978-3-642-03915-7_22}. The decision tree is built with an incremental design. In contrast to conventional decision trees, Hoeffding trees does not need to store samples in memory. Each node in the tree contains enough information to facilitate tree expansion and conduct regression. HATR incorporates an ADWIN \cite{bifet2009improving} concept-drift detector instance at each decision node to continually monitor potential changes in the data distribution. In the event of detecting a drift in a specific node, HATR initiates the induction of an alternative tree in the background. As sufficient information accumulates, HATR seamlessly replaces the node where the change was identified with its corresponding node from the alternate tree. This dynamic adaptation mechanism allows HATR to effectively handle concept drift and maintain accurate regression predictions in evolving data environments.

\subsubsection{Support Vector Regressor}

The Support Vector Regressor (SVR) technique, introduced by Cortes and Vapnik in 1995 within the realm of computer science \cite{cortes1995support}, involves mapping input vectors non-linearly into an expansive, high-dimensional feature space. Within this feature space, a linear decision surface is created. The distinctive characteristics of this decision surface ensure a strong generalization capability for the learning machine. SVRs can use different kernel functions (linear, polynomial, radial basis function, etc.) to handle various types of data and create non-linear decision boundaries. This flexibility allows SVRs to adapt to complex relationships within the data.

\subsubsection{Multi-layer Perceptron}

Multi-layer Perceptrons (MLPs) are intricate, massively parallel interconnected networks characterized by simple, often adaptive, elements and hierarchical organization \cite{white1992artificial}. These networks play a pivotal role in data analysis techniques, providing unparalleled flexibility for processing extensive volumes of multivariate, non-linear data. Unlike their more rigid and complex counterparts, artificial neural networks excel in adapting to intricate patterns and relationships within data, making them highly effective in tackling the challenges posed by large and complex datasets. Their capacity for learning and hierarchical representation enhances their utility in capturing nuanced features and optimizing performance across a diverse range of applications.

\subsubsection{Random Forest}

The Random Forest (RF) is an ensemble method that constructs multiple decision trees to collectively assess a new instance, ultimately determining its ranking through a majority vote \cite{breiman2001random}. Notably, each node in these decision trees employs a randomly selected subset of features from the original feature set. Moreover, each tree utilizes a distinct bootstrap data sample, akin to bagging techniques.

Bagging methods, characterized by the combination of multiple models, typically surpass the accuracy of individual classifiers. In contrast, boosting methods have the potential for even greater accuracy than bagging, but they tend to be more susceptible to noise. Random Forest stands out as a robust alternative, exhibiting resilience to noise while delivering comparable or superior performance to boosting methods. Importantly, Random Forest demonstrates this effectiveness without succumbing to overfitting, making it a versatile and reliable choice in various machine learning applications.

\subsubsection{DoME}

DoME is an algorithm that allows regression problems to be solved by developing an equation that represents the relationship between the independent variables and the dependent variable \cite{RIVERO2022116712}. This type of regression, in which the model sought is an equation, is called Symbolic Regression.

In the case of DoME, to do this the equations are represented internally with the shape of a tree, with a series of nodes, either terminal and non-terminal. As terminal nodes we will have the independent variables and different constants, and as non-terminal nodes we will have the operators that will be part of the equation. In this algorithm, the operators used are the arithmetic operators (+, -, *, /). In this way, the complexity of the resulting expressions can be bounded by placing a limit on the number of nodes that the tree can use.

Starting from an initial tree with a constant equal to the average value of the dependent variable, an iterative process is developed, in which, in each cycle, the tree is gradually modified until it reaches a point where no modification is found that leads to an improvement in the predictions.

\subsection{Feature extraction}

Principal Component Analysis (PCA) is a statistical technique used in the exploratory analysis of data \cite{MACKIEWICZ1993303}. One of its key applications involves reducing dimensionality by retaining as much information (variance) as possible. In scenarios with a large set of potentially correlated quantitative variables indicating redundant information, PCA enables the transformation of these variables into a reduced set of new variables known as principal components. These components effectively capture a significant portion of the data's variability. Each principal component is a linear combination of the original variables and is independent or uncorrelated with others. These derived principal components can then be employed in supervised learning approaches, such as principal component regression.

\subsection{Performance measures}

In the examination of the trained models and for subsequent comparisons, we employ 3 metrics designed specifically for regression problems. $R^2$ (coefficient of determination), MAE (mean absolute error) and RMSE (root mean squared error).

The $R^2$ score evaluates the effectiveness of the model by assessing its ability to elucidate variations in the data, prioritising this over quantifying the absolute number of accurate predictions. A key advantage of the $R^2$ score lies in its context independence. Consequently, the use of $R^2$ provides a foundational model for comparison, a feature not offered by other metrics. Essentially, $R^2$ squared measures the extent to which the regression line improves performance compared to a simple mean line. Calculated according to Eq. \ref{eq:r2_score}.

\begin{equation}
   R^2 = 1 - \frac{SSr}{SSm}
   \label{eq:r2_score}
\end{equation}

Where $SSr$ is the squared sum error of regression line and $SSm$ is the squared sum error of the mean of the observed data.

Computed using the formula in Eq. \ref{eq:mae}, MAE serves as a metric that quantifies the accuracy of the prediction between predicted and actual values.

\begin{equation}
   MAE = \frac{1}{n} \sum \left |  y - \hat{y} \right |
   \label{eq:mae}
\end{equation}

Where $n$ is the total number of data points, $y$ is the actual value and $\hat{y}$ is the predicted value.

Computed using Eq. \ref{eq:rmse}, RMSE is the square root of the mean squared error (MSE). MSE consists of calculating the squared difference between the actual and predicted values. Unlike MSE, the value resulting from calculating RMSE is in the same unit as the desired output variable, facilitating the interpretation of loss.

\begin{equation}
   RMSE = \sqrt{\frac{1}{n} \sum (y - \hat{y})^2}
   \label{eq:rmse}
\end{equation}

Where $n$ is the total number of data points, $y$ is the actual value and $\hat{y}$ is the predicted value.

\subsection{Experimentation setup}


We have studied the performance of the models by applying a Principal Component Analysis (PCA) to reduce the dimensionality of the dataset. The number of components are chosen internally by the method until reaching a desired explained variance. In this case, a explained variance of 99.9\% was chosen. Table \ref{tab:features} shows the difference in the number of input features used depending on whether we apply feature extraction or not.

\begin{table}[ht]
\centering
\resizebox{0.7\textwidth}{!}{%
\begin{tabular}{@{}ccccccc@{}}
\toprule
\rowcolor[HTML]{ECF4FF} 
 & \multicolumn{6}{c}{Input features} \\
\rowcolor[HTML]{ECF4FF}
\multirow{-2}{*}{Feature extraction} & A8 & A3 & P4 & P2 & V1 & V4\\
\midrule
No PCA & 1072 & 652 & 988 & 820 & 1184 & 594 \\
PCA & 16 & 13 & 15 & 15 & 16 & 14 \\
\bottomrule
\end{tabular}
}
\caption{This table shows the number of input features used for model training as a function of the station and whether we apply feature extraction or not.}
\label{tab:features}
\end{table}

To compare the two learning paradigms studied, Batch and Stream learning, we created two sets of models. The training of the models was applied with a grid search strategy for hyperparameter optimisation. The parameters tested can be seen in the table \ref{tab:configuration}. The first set consisted of HTR, HATR and kNN. This set was used to test the performance of Stream Learning. The second set of models, consisting of SVR, kNN, MLP, RF and DoME, was intended to test the Batch Learning paradigm. Since the kNN algorithm has been tested in both Stream Learning and Batch Learning forms, we will refer to them as kNN-SL and kNN-BL respectively.

\begin{table}[ht]
\centering
\resizebox{1\textwidth}{!}{%
\begin{tabular}{@{}lc@{}}
\toprule
\rowcolor[HTML]{ECF4FF} 
\textbf{k-Nearest Neighbors with Batch Learning} &                                  \\ \midrule
k value & 1, 3, 5, 7, 9 and 11 \\ \midrule
\rowcolor[HTML]{ECF4FF}
\textbf{k-Nearest Neighbors with Stream Learning} &                                  \\ \midrule
k value                            & 1, 3, 5, 7, 9 and 11                               \\
Maximum number of neighbors per node & 20                               \\
Maximum buffer size & 1000                               \\
\midrule \rowcolor[HTML]{ECF4FF} 
\textbf{Hoeffding Tree Regressor} &                                  \\ \midrule
Grace period & 7, 14, 30, 180 and 365 \\
Delta &  $1e^{-5}$, $1e^{-6}$, $1e^{-7}$ and $1e^{-8}$ \\
Model selector decay & 0.1, 0.2, 0.3, 0.4, 0.5, 0.6, 0.7, 0.8, 0.9 and 1 \\
Tau & 0.01, 0.05 and 0.1 \\
\midrule \rowcolor[HTML]{ECF4FF} 
\textbf{Hoeffding Adaptative Tree Regressor} &                                  \\ \midrule
Grace period & 7, 14, 30, 180 and 365 \\
Delta &  $1e^{-5}$, $1e^{-6}$, $1e^{-7}$ and $1e^{-8}$  \\
Model selector decay & 0.1, 0.2, 0.3, 0.4, 0.5, 0.6, 0.7, 0.8, 0.9 and 1 \\
Tau & 0.01, 0.05 and 0.1 \\
\midrule \rowcolor[HTML]{ECF4FF} 
\textbf{Support Vector Regressor} &                                  \\ \midrule
Kernel type            & Lineal, Gaussian and Polynomial \\ 
C value                & 0.001, 0.01, 0.05, 0.1, 1 and 10                              \\ 
Epsilon        & 0.1, 0.2, 0.3, 0.4, 0.5, 0.6, 0.7, 0.8, 0.9 and 1            \\ 
Grade (polynomial kernel) & 1, 2 and 3                           \\ 
\midrule \rowcolor[HTML]{ECF4FF} 
\textbf{Multi-Layer Perceptron} &                                  \\ \midrule
Number of neurons in a  & \multirow{2}{*}{2, 4, 8 and 10}                \\
\ one hidden layer network & \\
Number of neurons in a  & {(}2, 2{)}, {(}4, 2{)}, {(}10, 2{)}, {(}10, 10{)},  \\                    
\ two hidden layers network & {(}16, 8{)}, {(}32, 8{)}, {(}32, 16{)}\\ 
Activation function output layer        & Sigmoid                 \\ 
Hidden layers activation function      & Relu                     \\ 
Optimizer                              & RMSprop                     \\ 
Learning rate                            & $1e^{-1}$, $1e^{-2}$, $1e^{-3}$, $1e^{-4}$ and $1e^{-5}$                    \\ 
Number of epochs                         & 500                       \\
\midrule \rowcolor[HTML]{ECF4FF} 
\textbf{Random Forest} &                                  \\ \midrule
Quality criterion of a split & Squared error, Friedman mse and Poisson \\
Max depth & 2, 4, 10, 20, 50 and maximum \\
Number of trees        & 1000   \\
\midrule \rowcolor[HTML]{ECF4FF} 
\textbf{DoME} &                                  \\ \midrule
Minimum Reductions MSE & $1e^{-1}$, $1e^{-2}$, $1e^{-3}$, $1e^{-4}$, $1e^{-5}$, $1e^{-6}$ and $1e^{-7}$  \\
MaxNumNodes & 5:5:100\\
Strategy & Selective with constant optimization\\
\bottomrule
\end{tabular}%
}
\caption{Summary table of the models parameter values used in the grid search.}
\label{tab:configuration}
\end{table}

For the validation process of the results, we have followed a hold-out strategy. Of the 7 years of data available (2013 - 2019), the last year (2019) has been reserved to validate the models. This is because it is expected that the developed model can be put into production and the most realistic test, being a time series, is to train with past experience and test with the latest data. 

\begin{figure}[ht]
  \centering
  \includegraphics[width=1\textwidth]{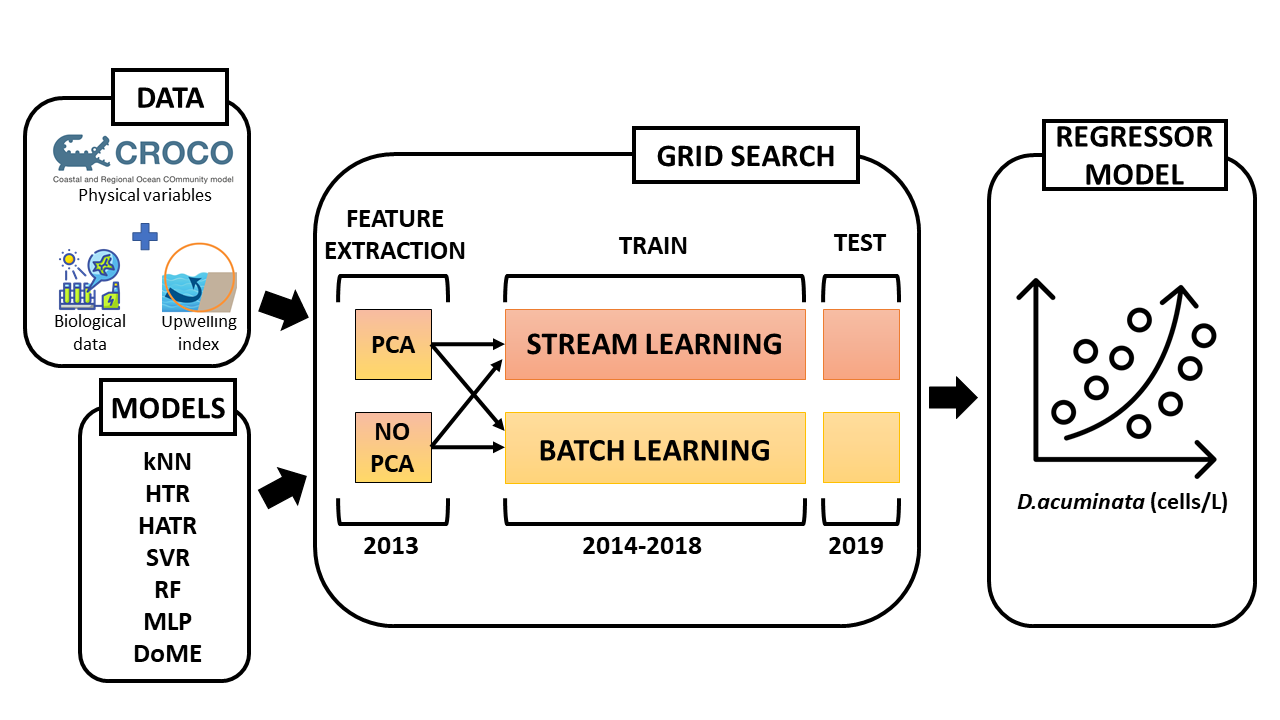}
  \caption{Schematic representation of the machine learning-based system proposed for HAB forecasting.}
  \label{fig:workflow}
\end{figure}

The first year of data (2013) was used to perform a pre-training of the models. This pre-training makes it possible to apply the PCA algorithm in a way that does not affect the comparison between the learning paradigms. This is necessary because PCA is applied statically, and setting up feature extraction with the entire training set would go against the Stream Learning principle of training the model sequentially and dynamically. Once the regression models and PCA are pre-trained, we proceed to train the models based on the applied paradigm and finish with a validation process with data never seen by the models. This process allows us to make an unbiased comparison between the various models and paradigms. The workflow shown in figure \ref{fig:workflow} allows us to see more clearly the different steps carried out during the training and comparison of the techniques studied.

\section{Results}
\label{sec:r}

In this study, a comparison of models was carried out in order to find the one that best adapts to forecast the concentration of \textit{Dynophysis acuminata} (cells/L) cells. The prediction was performed at six mooring stations, which were selected as the target locations, representative of the inner and outer Ría. For this comparison we tested 8 machine learning models and the use of a dimensionality reduction technique (PCA), having a total of 16 combinations. These were tested for each predicted day, 1, 2, 3, 4, 5, 6 and 7 days at each predicted station. Table \ref{tab:rank} shows a ranking of the 16 combinations for each forecast day, which were ordered by their R$^{2}$ values averaged for all the predicted stations. Based on that we assigned 1 point to the first and 16 to the last (Rank per prediction day in Table \ref{tab:rank}). The average rank of the 16 combinations for all forecast days (last column in table \ref{tab:rank}), was calculated by averaging the rank of each model (each row in Table \ref{tab:rank}). This table shows certain trends in the results. Models that internally incorporate feature selection methods like RF or DoME seem to perform better without applying PCA. On the other side, the other methods perform better with a reduced set of features.  We can also observe that up to 5 days of prediction the different combinations remain in the same positions in the ranking, while in the last days of prediction studied there is a high variability as to which model offers the best results.

\begin{table}[ht]
\resizebox{1\textwidth}{!}{%
\begin{tabular}{@{}ccccccccccc@{}}
\toprule
\rowcolor[HTML]{ECF4FF} 
OVERALL &  & FEATURE & \multicolumn{8}{c}{RANK PER PREDICTION DAY} \\
\rowcolor[HTML]{ECF4FF} 
RANK & \multirow{-2}{*}{MODEL} & EXTRACTION & 1 & 2 & 3 & 4 & 5 & 6 & 7 & AVERAGE \\ \midrule
1 & DoME & No PCA & 1 & 1 & 1 & 1 & 2 & 2 & 10 & 2,57 \\
2 & SVR & PCA & 4 & 4 & 4 & 4 & 3 & 1 & 3 & 3,29 \\
3 & DoME & PCA & 3 & 3 & 3 & 2 & 1 & 3 & 9 & 3,43 \\ \rowcolor[HTML]{E4E4E4}
4 & HATR & PCA & 5 & 5 & 5 & 5 & 5 & 5 & 2 & 4,57 \\
5 & RF & No PCA & 2 & 2 & 2 & 3 & 8 & 12 & 12 & 5,86 \\ \rowcolor[HTML]{E4E4E4}
5 & HTR & PCA & 7 & 6 & 6 & 6 & 4 & 7 & 5 & 5,86 \\
7 & SVR & No PCA & 10 & 9 & 9 & 7 & 7 & 6 & 1 & 7,00 \\
8 & MLP & PCA & 8 & 8 & 8 & 8 & 9 & 9 & 6 & 8,00 \\ \rowcolor[HTML]{E4E4E4}
9 & HATR & No PCA & 12 & 10 & 10 & 10 & 10 & 4 & 4 & 8,57 \\ \rowcolor[HTML]{E4E4E4}
10 & HTR & No PCA & 13 & 12 & 11 & 9 & 6 & 8 & 7 & 9,43 \\
11 & RF & PCA & 6 & 7 & 7 & 12 & 13 & 13 & 13 & 10,14 \\ \rowcolor[HTML]{E4E4E4}
12 & kNN-SL & PCA & 11 & 11 & 12 & 11 & 11 & 11 & 11 & 11,14 \\
13 & MLP & No PCA & 15 & 14 & 14 & 13 & 12 & 10 & 8 & 12,29 \\
14 & kNN-BL & PCA & 14 & 13 & 13 & 14 & 14 & 14 & 15 & 13,86 \\ \rowcolor[HTML]{E4E4E4}
15 & kNN-SL & No PCA & 9 & 15 & 15 & 15 & 15 & 15 & 14 & 14,00 \\
16 & kNN-BL & No PCA & 16 & 16 & 16 & 16 & 16 & 16 & 16 & 16,00 \\ \bottomrule
\end{tabular}
}
\caption{This ranking of the models studied shows the learning paradigm used and whether feature extraction by PCA has been applied. This ranking is also displayed according to the prediction days. Models trained with the Stream Learning paradigm are marked in grey while those trained with Batch Learning are marked in white. The values used for the rankings come from the average $R^2$ score obtained in the 6 oceanographic stations.}
\label{tab:rank}
\end{table}

\begin{figure}[ht]
  \centering
  \includegraphics[width=1\textwidth]{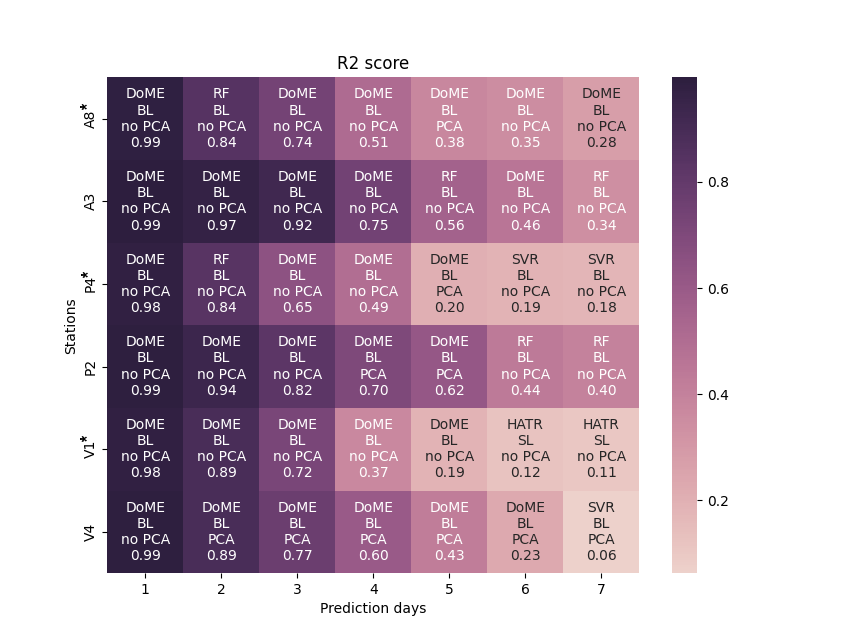}
  \caption{Heat map showing the $R^2$ score obtained by the best model for each predicted station, from 1 to 7 days of prediction. In addition, this map shows the model that obtained this result, the learning paradigm used and whether it was obtained with or without PCA. Outer stations are indicated with an asterisk.}
  \label{fig:heatMap}
\end{figure}

In figure \ref{fig:heatMap} we can observe the results offered by the proposed pipeline. More specifically, we can see which particular model gave the best result at each predicted station, the learning paradigm used and whether they were obtained after applying PCA or not. In this figure, we can see how applying PCA as a feature reducer works better for the V4 (inner) station, while not applying it offers better results for the rest. The difference in difficulty for each station can also be seen, with outer stations, A8, P4 and V1, being the most difficult. Another observation we can make is that the accuracy of the models decreases when predicting for 4 days or further. Although in inner stations at Ría de Arousa, A3, and Pontevedra, P2, it would be possible to forecast more days ahead while maintaining a certain quality in the predictions. We have defined the threshold at 3 days for the following comparisons. 3-day-ahead is a critical point in the predictions, because a model with this capacity could provide support for holidays and weekends, as samples are not collected during this days.

\begin{figure}[ht]
  \centering
  \includegraphics[width=1\textwidth]{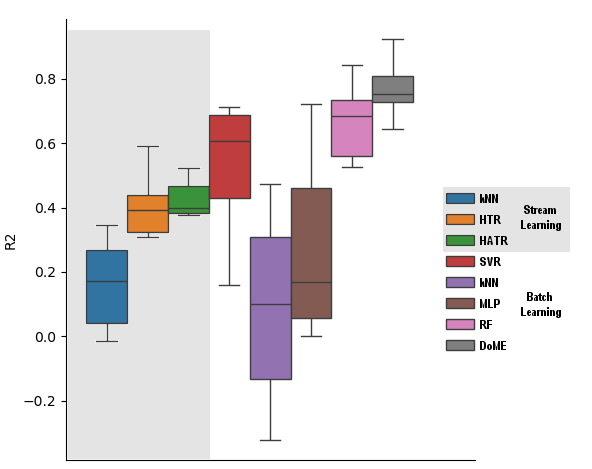}
  \caption{Box plot showing the results obtained by applying the models to predict at six monitoring stations and 3-days ahead.}
  \label{fig:boxPlot}
\end{figure}

Figure \ref{fig:boxPlot} shows the behaviour of the models studied when predicting the concentration of \textit{D. acuminata} at 3 days in the future, while in the table \ref{tab:resultados_resumen} these results are shown in a more detailed way, as well as the values of MAE and RMSE. For each model, the results offered in each of the 6 forecasted stations are shown, so we can observe the adequacy of the models and learning paradigms for the abundance prediction of \textit{D. acuminata} cells in a global way. The DoME model obtains the best results in all the metrics, followed by RF model. Although, with a standard deviation of $\pm$ 0.12 for the $R^2$ of RF it may seem that DoME is not always the best. however, if we look at the table \ref{tab:general_results} (located in the Appendices) we can see that these deviations are due to the station where the model is applied and that in any of the cases it is DoME that offers the best metrics. Regarding the models trained with the Stream Learning paradigm, the HTR model and its variant HATR offer very similar results overall. However, looking at the table \ref{tab:general_results} (in Appendices) showing the results divided by monitoring station, we can see that the model that best adapts varies depending on the area where it is applied. HTR is the best in A8 (outer), P2 and V4 (inner stations) while HATR is the best in A3 (inner), P4 and V1 (outer stations).

\begin{table}[ht]
\resizebox{1\textwidth}{!}{%
\begin{tabular}{@{}clllllll@{}}
\toprule
\rowcolor[HTML]{ECF4FF} 
PARADIGM &  & \multicolumn{2}{c}{} & \multicolumn{2}{c}{} & \multicolumn{2}{c}{} \\
\rowcolor[HTML]{ECF4FF} 
LEARNING & \multirow{-2}{*}{MODEL} & \multicolumn{2}{c}{\multirow{-2}{*}{$R^2$}} & \multicolumn{2}{c}{\multirow{-2}{*}{MAE}} & \multicolumn{2}{c}{\multirow{-2}{*}{RMSE}} 
\\ \midrule
\multirow{3}{*}{Stream Learning} & kNN-SL & 0.17 & $\pm$ 0.15 & 167.80 & $\pm$ 141.51 & 357.72 & $\pm$ 370.23 \\
 & HTR & 0.41 & $\pm$ 0.11 & 143.16 & $\pm$ 115.12 & 285.98 & $\pm$ 272.99 \\
 & HATR & 0.43 & $\pm$ 0.06 & 147.18 & $\pm$ 147.00 & 297.53 & $\pm$ 311.26 \\ \midrule
\multirow{5}{*}{Batch Learning} & SVR & 0.53 & $\pm$ 0.22 & 111.18 & $\pm$ 92.09 & 255.51 & $\pm$ 233.27 \\
 & kNN-BL & 0.09 & $\pm$ 0.31 & 172.91 & $\pm$ 134.24 & 349.86 & $\pm$ 308.37 \\
 & MLP & 0.27 & $\pm$ 0.29 & 171.99 & $\pm$ 123.25 & 289.08 & $\pm$ 218.67 \\
 & RF & 0.67 & $\pm$ 0.12 & 100.21 & $\pm$ 78.41 & 225.80 & $\pm$ 221.05 \\
 & \textbf{DoME} & \textbf{0.77} & \textbf{$\mathbf{\pm}$ 0.09} & \textbf{87.78} & \textbf{$\mathbf{\pm}$ 69.25} & \textbf{187.57} & \textbf{$\mathbf{\pm}$ 182.78} \\ \bottomrule
\end{tabular}
}
\caption{Summary table with the metrics of each model averaged in the six monitoring stations, where predictions were made, and 3-days ahead.}
\label{tab:resultados_resumen}
\end{table}

\begin{figure}[ht]
  \centering
  \includegraphics[width=1\textwidth]{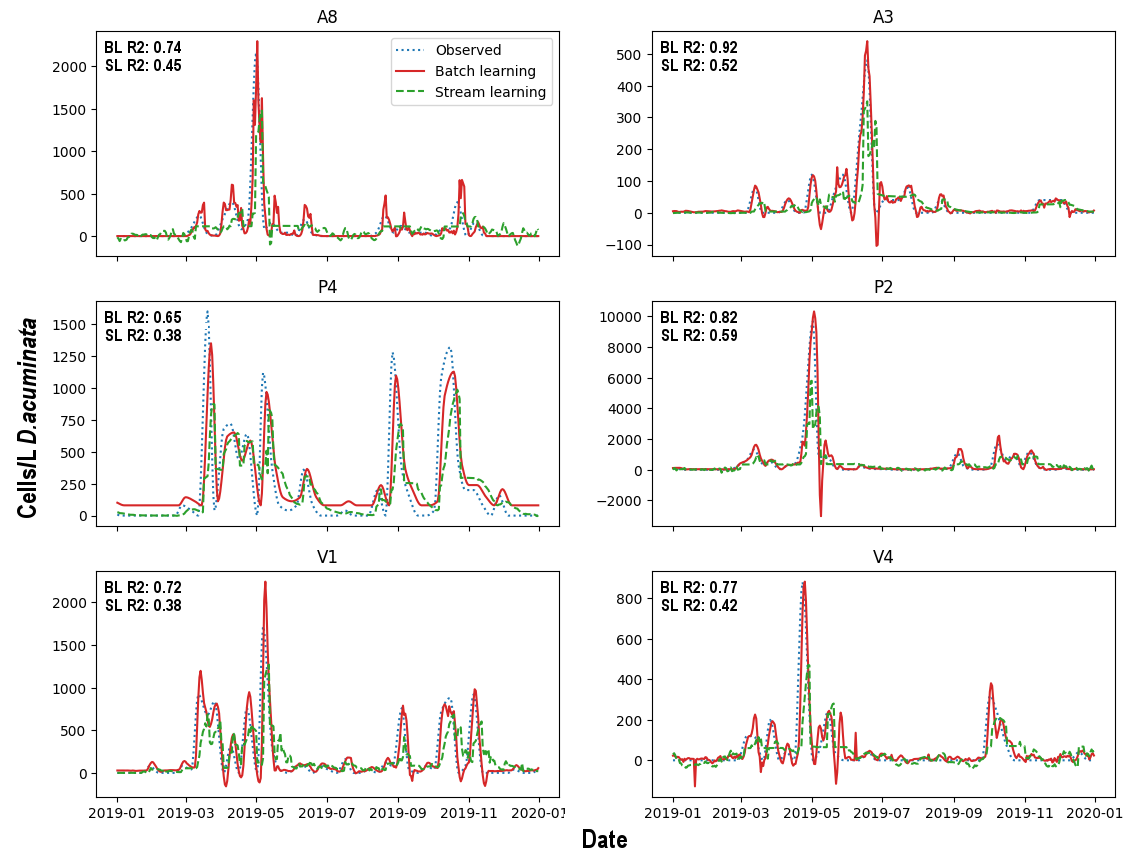}
  \caption{Graphs with the concentration values of observed \textit{D. acuminata} and predicted by DoME and HTR/HATR (HTR at A8, P2 and V4; HATR at A3, P4 and V1). These models give the best results for BL and SL respectively. The predictions made in these graphs are made 3 days in advance of the test period.}
  \label{fig:BlvsSl}
\end{figure}

One of the objectives of this study was to test the applicability of Stream Learning in HAB prediction. Therefore, in the Figure \ref{fig:BlvsSl} we can observe the results of the best model for each paradigm and how it fits the test data (2019). HTR or HATR are shown for Stream Learning (HTR in A8, P2 and V4; HATR in A3, P4 and V1) and DoME for Batch Learning. 

The DoME model has one main advantage and that is the high explanatory power it offers. Once trained, this model results in an equation with the most relevant features for the prediction. An example of this can be seen in the equation \ref{eq:A8_eq} belonging to the model created to make 3-day predictions at the outer mooring station, A8, at Ría de Arousa. In this equation we can see how the prediction of \textit{D. acuminata} is calculated based on the current concentration, the concentration 4 days ago and the standard deviation of the meridional (v) component of the sea current.

\begin{equation}
   \resizebox{1\hsize}{!}{
   $DacumA8_{i+3} = 6.2309 + DacumA8_{i}\cdot \left(0.2647+ \frac{4.5940 \cdot e^{-3} - 4.7374\cdot e^{-7}\cdot DacumA8_{i-4}}{stdVbottomSA3_{i-1}}\right)$
   }
   \label{eq:A8_eq}
\end{equation}

Where $DacumA8_{i}$ is the concentration of \textit{D. acuminata} collected at outer station, A8, at the current day (i), $DacumA8_{i-4}$ the same but four days ago (i-4) and $stdVbottomSA3_{i-1}$ is the standard deviation of the meridional (v) velocity component at the bottom layer of the first half of SA3 section one day ago. The first half of SA3 section (black line in Figure \ref{fig:mapa}), at Ría de Aoursa, is defined from the black square to the cross-line.



\section{Discussion} 
\label{sec:d}

The main objective of this study was to create a methodology that fairly compares a series of models trained with two learning paradigms. This methodology had to provide the best model configured with the optimal hyperparameters. The result of this process can be seen in figure \ref{fig:heatMap}, where this methodology is applied to the six forecast monitoring stations and with predictions from 1 to 7 days in the future. A local approach was used to develop prediction models for the concentration of \textit{D. acuminata} cells, where each predicted monitoring station studied had its own models. This allowed us to observe the robustness of the predictions by testing each configuration in 6 different locations, representative of the outer and inner Ría de Arousa, Pontevedra and Vigo. We can see how stations A8, P4 and V1 located in the outer Rías are more difficult to predict compared to stations located in the inner Rías, such as A3, P2 and V4. While in our models the inner predicted stations contain information from the surrounding stations and sections, the lack of this information may be occurring with the prediction at the outer stations, which are strongly linked to what happens on the shelf. As noted by some authors, HAB events can be initiated in the Galician Rias Baixas by the advection of shelf populations \cite{Moita2016, Diaz2019a, Hariri2022} from southern locations, transported by currents \cite{Sordo2001, Peliz2002}. HAB initiation by advection of allochthonous populations has also been described elsewhere in the world, such as in Chile (i.e. \cite{Rosales2024}), the Gulf of Maine (i.e. \cite{Li2014}), Scottish coastal waters (i.e. \cite{Gillibrand2016}) and between different regions at the European Atlantic shelf (i.e. \cite{Bedington2022}). Despite these regional characteristics of the Rías Baixas, we can observe that it is feasible to develop predictive models of HABs adapted to heterogeneous environments trained with output data from the hydrodynamic model, CROCO. Future works on HABs prediction, using machine learning techniques, could explore the addition of information from some shelf stations, emphasizing the necessity for monitoring programs at shelf locations to anticipate the advection of HABs into the Galician Rias Baixas.

In relation to the days that can be predicted with some degree of confidence, we set 3 days in advance as the threshold, where R$^{2}$ higher than 0.6 can be found (Figure \ref{fig:heatMap}). This limit in the prediction can be related with sudden changes in the oceanographic conditions of the Rias, where the upwelling and downwelling events have a duration close to these number of days \cite{Gilcoto2017}. Although, at the inner stations of Ría de Arousa, A3, and Pontevedra, P2, the models continue to offer interesting results even up to 5 days in the future. When predicting 4 days ahead or further, the $R^2$ scores decay below 0.5 in some of the stations studied, showing that these models are not good enough. 

From all the models testes, DoME, with an $R^2$ score of 0.77 in 3-day prediction, is the one that offers the best results, followed by RF with 0.67 and SVR with 0.53. These models offer better results than MLP, one of the most widely used models in the prediction of HAB \cite{1553742, GUO2020111731}. In particular, we can see how the configuration used in another region \cite{VeloSurez2007ArtificialNN}, also on the Spanish coast (Huelva), does not work for the geography of the Galician coast. Furthermore, as can be seen in the example equation \ref{eq:A8_eq}, once trained, the DoME model can be represented in equation format that can be applied directly to the features. This offers a very fast and easy-to-apply tool for HAB prediction for the aquaculture industry. This great advantage of the models with high explainability has not been sufficiently studied in this field. Models such as Extra Tree Regressor or Random Forest \cite{YAN2024169253}, are more frequently used for their explainability, but, with a 0.44 $R^2$ in test, offer worse results than the proposed model.

Once the 3-day prediction space was determined, we compared the Batch Learning and Stream Learning paradigms. We sought to see if the paradigm specialised in time series and their fluctuations was better adapted to describe the variability of \textit{D. acuminata} cells of the last few years \cite{BOIVINRIOUX2022102183}. In figure \ref{fig:BlvsSl} we can see that in none of the studied scenarios this is the case. We can therefore conclude that these anomalies are yet not so decisive in the face of the strong seasonal component of the biophysical variables used as input. 

In this study, the hydrodynamic CROCO model was used in hindcast mode to provide the variables needed to train and test the machine learning models. However, CROCO can also be run in the forecasting mode, to predict oceanographic conditions a few days in advance. Knowledge of the conditions that favour the initiation, growth and transport of harmful algal blooms, together with these tools, can help to obtain an accurate prediction.


\section{Conclusions} 
\label{sec:c}

In this research, we established a framework for forecasting \textit{D. acuminata} concentrations, along the Galician coast (Spain). We employed 7 machine learning algorithms, encompassing two training paradigms. Our findings indicate that the DoME algorithm without a previous selection of features through PCA, outperformed all others as the most effective predictor of \textit{D. acuminata} concentrations. Consequently, we highly advocate for the adoption of this model in future studies focusing on predicting \textit{D. acuminata} concentrations. Considering the comparison between paradigms, the results suggest that the years 2013-2019 do not show enough fluctuations in the data for Stream Learning to outperform Batch Learning. 

Faced with a limited amount of data, we have demonstrated how the use of daily oceanographic variables obtained from hydrodynamic models, such as CROCO, can offer new opportunities in the prediction of algae blooms. In particular, it has allowed us to make daily predictions without the need to resort to weekly predictions.

It would be interesting to use 3D hydrodynamic models, like CROCO, as a predictor of physical variables in combination with the proposed model as a predictor of \textit{D. acuminata} concentrations, to create a dataset to feed a new model capable of accurately predicting more days into the future.

Recognizing the importance of a system capable of monitoring and predicting Harmful Algal Bloom (HAB) events is crucial. Inadequate preparation for these natural occurrences can result in substantial economic losses for aquaculture and shellfish industries. Therefore, we believe that models such as the one proposed here can help producers to improve resource management by anticipating possible closure events in production areas within a few days, and to optimize the resources of the monitoring service in the event of possible HABs episodes. Notably, emphasizing factors such as developing highly interpretable models and training them with data obtained from hydrodynamic models, like CROCO, to compensate for limited sampled data is essential.
It is in these aspects that our work distinguishes itself, enhancing results beyond those achieved in prior studies.



\section*{Acknowledgments}
The authors want to acknowledge the support from INTECMAR, who has provided part of the data for this work; and CESGA, who allowed the conduction of the tests in their installations. Thanks are also due for the financial support to CESAM by FCT/MCTES (UIDP/50017/2020+UIDB/ 50017/2020+LA/P/0094/2020), through national funds, and the co- funding by the FEDER, within the PT2020 Partnership Agreement and Compete 2020. Funding for open access charge: Universidade da Coruña/CISUG. CITIC is funded by the Xunta de Galicia through the collaboration agreement between the Regional Ministry of Culture, Education, Vocational Training and Universities and the Galician universities to strengthen the research centres of the Galician University System (CIGUS). Grant PID2021-126289OA-I00 funded by MCIN/AEI/10.13039/501100011033 and by ERDF A way of making Europe. Elisabet R. Cruz was supported by the Portuguese Science and Technology Foundation (FCT) through PhD fellowship PD/BD/143085/2018. This research was funded by REDEIRA (Research, development and innovation of a Coastal Observation network: Ría de Arousa) project (Proyecto Estratégico Orientado a la Transición Ecológica y a la Transición Digital; Ref: TED2021-132188B-I00).

\section*{Data availability and reproducibility}
The source code of the analysis is available on GitHub: \url{https://github.com/AndresMolares/HAB-forecasting-SL-vs-BL}. Part of the dataset was obtained following an official request to INTECMAR which does not allow its redistribution.

\bibliographystyle{unsrt}  
\bibliography{references}  

\begin{appendices}

\begin{table}[ht]
\centering
\resizebox{0.9\textwidth}{!}{%
\begin{tabular}{@{}cclclll@{}}
\toprule
\rowcolor[HTML]{ECF4FF} 
 & LEARNING & & FEATURE &  &  &  \\ 
 \rowcolor[HTML]{ECF4FF} 
\multirow{-2}{*}{STATION} & PARADIGM & \multirow{-2}{*}{MODEL} & EXTRACTION & \multirow{-2}{*}{$R^2$} & \multirow{-2}{*}{MAE} & \multirow{-2}{*}{RMSE} \\ 
\midrule
\multirow{8}{*}{A8} & \multirow{3}{*}{Stream Learning} & kNN-SL & PCA & -0.01 & 112.03 & 268.96 \\
 &  & HTR & PCA & 0.45 & 93.26 & 198.33 \\
 &  & HATR & PCA & 0.41 & 85.14 & 205.01 \\
 \cmidrule(l){2-7} 
 & \multirow{5}{*}{Batch Learning} & SVR & PCA & 0.60 & 60.73 & 168.60 \\
 &  & kNN-BL & PCA & 0.07 & 116.48 & 258.12 \\
 &  & MLP & PCA & 0.09 & 110.78 & 254.95 \\
 &  & RF & PCA & 0.65 & 94.71 & 156.96 \\
 &  & \textbf{DoME} & \textbf{No PCA} & \textbf{0.74} & \textbf{58.35} & \textbf{136.35} \\
 \midrule
\multirow{8}{*}{A3} & \multirow{3}{*}{Stream Learning} & kNN-SL & PCA & 0.35 & 29.48 & 56.47 \\
 &  & HTR & PCA & 0.37 & 25.22 & 55.47 \\
 &  & HATR & PCA & 0.52 & 22.13 & 48.14 \\
 \cmidrule(l){2-7} 
 & \multirow{5}{*}{Batch Learning} & SVR & PCA & 0.61 & 19.95 & 43.31 \\
 &  & kNN-BL & PCA & 0.36 & 29.39 & 55.63 \\
 &  & MLP & PCA & 0.53 & 26.54 & 47.44 \\
 &  & RF & No PCA & 0.84 & 11.88 & 27.51 \\
 &  & \textbf{DoME} & \textbf{No PCA} & \textbf{0.92} & \textbf{11.67} & \textbf{19.44} \\
 \midrule
\multirow{8}{*}{P4} & \multirow{3}{*}{Stream Learning} & kNN-SL & PCA & 0.03 & 239.71 & 351.33 \\
 &  & HTR & PCA & 0.31 & 194.83 & 296.32 \\
 &  & HATR & PCA & 0.38 & 159.37 & 279.99 \\
 \cmidrule(l){2-7} 
 & \multirow{5}{*}{Batch Learning} & SVR & No PCA & 0.16 & 207.78 & 326.70 \\
 &  & kNN-BL & PCA & -0.20 & 248.31 & 389.63 \\
 &  & MLP & PCA & 0.00 & 266.26 & 355.76 \\
 &  & RF & No PCA & 0.53 & 123.97 & 243.85 \\
 &  & \textbf{DoME} & \textbf{No PCA} & \textbf{0.65} & \textbf{142.43} & \textbf{212.07} \\
 \midrule
\multirow{8}{*}{P2} & \multirow{3}{*}{Stream Learning} & kNN-SL & PCA & 0.27 & 411.52 & 1080.00 \\
 &  & HTR & PCA & 0.59 & 335.36 & 809.67 \\
 &  & HATR & No PCA & 0.49 & 426.13 & 906.30 \\
 \cmidrule(l){2-7} 
 & \multirow{5}{*}{Batch Learning} & SVR & PCA & 0.71 & 231.15 & 678.03 \\
 &  & kNN-BL & PCA & 0.47 & 387.90 & 918.43 \\
 &  & MLP & PCA & 0.72 & 344.81 & 664.24 \\
 &  & RF & No PCA & 0.74 & 228.68 & 643.31 \\
 &  & \textbf{DoME} & \textbf{No PCA} & \textbf{0.82} & \textbf{192.80} & \textbf{532.00} \\
 \midrule
\multirow{8}{*}{V1} & \multirow{3}{*}{Stream Learning} & kNN-SL & PCA & 0.26 & 161.45 & 271.95 \\
 &  & HTR & PCA & 0.31 & 165.82 & 262.16 \\
 &  & HATR & PCA & 0.38 & 146.05 & 249.21 \\
 \cmidrule(l){2-7} 
 & \multirow{5}{*}{Batch Learning} & SVR & PCA & 0.37 & 123.79 & 250.33 \\
 &  & kNN-BL & PCA & -0.32 & 199.54 & 362.94 \\
 &  & MLP & PCA & 0.05 & 214.13 & 307.38 \\
 &  & RF & No PCA & 0.53 & 116.48 & 217.19 \\
 &  & \textbf{DoME} & \textbf{No PCA} & \textbf{0.72} & \textbf{91.04} & \textbf{166.32} \\
 \midrule
\multirow{8}{*}{V4} & \multirow{3}{*}{Stream Learning} & kNN-SL & PCA & 0.09 & 52.61 & 117.63 \\
 &  & HTR & PCA & 0.42 & 44.48 & 93.94 \\
 &  & HATR & PCA & 0.38 & 44.23 & 96.53 \\
 \cmidrule(l){2-7} 
 & \multirow{5}{*}{Batch Learning} & SVR & PCA & 0.71 & 23.70 & 66.09 \\
 &  & kNN-BL & PCA & 0.14 & 55.82 & 114.41 \\
 &  & MLP & PCA & 0.25 & 69.44 & 104.71 \\
 &  & RF & No PCA & 0.71 & 25.56 & 66.00 \\
 &  & \textbf{DoME} & \textbf{PCA} & \textbf{0.77} & \textbf{30.41} & \textbf{59.21} \\ 
 \bottomrule
\end{tabular}
}
\caption{Best 3-day ahead \textit{D. acuminata} prediction results. These results are broken down by model and by oceanographic station where they were trained. The results shown in bold show the best model for each monitoring station.}
\label{tab:general_results}
\end{table}

\end{appendices}

\end{document}